\documentclass[final]{article}

\PassOptionsToPackage{numbers}{natbib}
\usepackage[preprint]{neurips_2019}

\usepackage[utf8]{inputenc} % allow utf-8 input
\usepackage[T1]{fontenc}    % use 8-bit T1 fonts
\usepackage{hyperref}       % hyperlinks
\usepackage{url}            % simple URL typesetting
\usepackage{booktabs}       % professional-quality tables
\usepackage{amsfonts}       % blackboard math symbols
\usepackage{nicefrac}       % compact symbols for 1/2, etc.
\usepackage{microtype}      % microtypography
\usepackage[dvipsnames]{xcolor}%

\usepackage[pdftex]{graphicx}
\usepackage{pifont}
\usepackage[frozencache,cachedir=.]{minted}%
\usepackage{authblk}%
\usepackage{caption}%
\usepackage{subcaption}%

\newcommand{\TorchBeast}{TorchBeast}%
\newcommand{\Cpp}{C\kern-.03em\raise.16ex\hbox{\small{+\kern-.03em+}}}%
\newcommand{\UNIX}{\textsc{unix}}%

\definecolor{red}{HTML}{f7b6d2}
\definecolor{blue}{HTML}{aec7e8}

\title{TorchBeast: A PyTorch Platform for Distributed RL}

\author[1]{Heinrich Küttler\thanks{Correspondence to hnr@fb.com.}}
\author[1,2]{Nantas Nardelli}
\author[1]{Thibaut Lavril}
\author[1,3]{Marco Selvatici}
\author[1]{Viswanath Sivakumar}
\author[1,4]{Tim Rocktäschel}
\author[1,4]{Edward Grefenstette}

\affil[1]{Facebook AI Research}
\affil[2]{University of Oxford}
\affil[3]{Imperial College London}
\affil[4]{University College London}

\begin{document}
\frenchspacing
\maketitle

\begin{abstract}
TorchBeast is a platform for reinforcement learning (RL)
research in PyTorch.  It implements a version of the popular IMPALA
algorithm~\cite{DBLP:journals/corr/abs-1802-01561} for fast,
asynchronous, parallel training of RL agents. Additionally, TorchBeast
has simplicity as an explicit design goal: We provide both a
pure-Python implementation (``MonoBeast'') as well as a multi-machine
high-performance version (``PolyBeast'').
In the latter, parts of the implementation are written in \Cpp, but all parts pertaining to machine learning are kept in simple Python using
PyTorch~\cite{paszke2017automatic}, with the environments provided
using the OpenAI Gym interface~\cite{1606.01540}.
This enables researchers to conduct scalable RL research using TorchBeast without any programming knowledge beyond Python and PyTorch.
In this paper, we describe the TorchBeast design principles and implementation and demonstrate that it performs on-par with IMPALA on Atari.
TorchBeast is released as an open-source package under the Apache 2.0 license and is available at \url{https://github.com/facebookresearch/torchbeast}.
\end{abstract}

\section{Introduction}

Reinforcement learning has recently had a surge of interest thanks to
the rise of deep learning and new GPU hardware, conquering
important challenges such as chess, Go, and other board
games~\citep{go, chessshogi}, demonstrating the ability to learn
policies on visual inputs~\citep{mnih2015human, levinerobot}, and tackle
strategically complex environments~\citep{gehring2018high, alphastarblog, OpenAI_dota}
as well as multi-agent settings~\citep{foerster2017stabilising, lowe2017multi}.
However a lack of well-written, high-performance, scalable
implementations of distributed RL architectures has hindered the
reproduction of published work, and largely restricted the development
of new work to a few organizations with the required know-how.
For model-free reinforcement learning with discrete action spaces,
approaches built on top of the IMPALA
agent~\cite{DBLP:journals/corr/abs-1802-01561} have achieved
prominence for domains like StarCraft II~\cite{alphastarblog} or
first-person shooter games~\cite{Jaderberg859}.
While an authoritative implementation of the IMPALA agent built on
TensorFlow~\cite{tensorflow2015-whitepaper} has been released
as open source software, researchers preferring PyTorch had fewer options.
TorchBeast aims to help leveling the playing field by being a simple
and readable PyTorch implementation of IMPALA, designed from the
ground-up to be easy-to-use, scalable, and fast.

\section{Background}
We consider the case of a Markov Decision Process (MDP) with a state
space $\mathcal{S}$, a set of actions $\mathcal{A}$ and a transition
function $\mathcal{T}: \mathcal{S}\times\mathcal{A}\to
\mathcal{P}(\mathcal{S})$ specifying the probability distribution over
next states given a state and action.
The agent receives rewards $r:
\mathcal{S}\times\mathcal{A}\to\mathbb{R}$ and attempts to maximize
the cumulative expected return within one episode $R_t = \mathbb{E}
\bigl[ \sum_{k = 0}^{T} \gamma^k r_{t+k+1} \bigr]$ for a discount
factor $0 < \gamma \le 1$.

IMPALA~\citep{DBLP:journals/corr/abs-1802-01561} uses an actor-critic
variant to update parameters $\omega$ of a policy $\pi_\omega$ as well
as a value-function estimate $V_\theta$ with parameters $\theta$.
Importantly, IMPALA is an off-policy method, maintaining a behavior
policy that acts in the environment and collects traces of experience,
as well as the current policy that we aim to update.
This update is achieved via calculating the policy gradient using an
importance weight between the behavior and current policy, thereby
correcting for the off-policiness of the behavior policy with respect
to the current policy. The specific off-policy correction method used
by IMPALA, V-trace, is more stable than to other such methods for
actor-critic agents. We refer to
\citep{DBLP:journals/corr/abs-1802-01561} for details).
% $\mathbb{E} \left[ \log\pi_\omega(a_t|s_t)Q(s_t,a_t)\right]$

\paragraph{Actors, learner and rollouts.} The IMPALA architecture
consists of a single \emph{learner} and several \emph{actors}. Each
actor produces \emph{rollouts} in an indefinite loop. A rollout
consists of \texttt{unroll\_length} many environment-agent
interactions. The learner then consumes batches of these rollouts. A
typical learner input might be a Python dictionary of the form
\begin{minted}{python}
{
  "observation": tensor(T, B, *obs_shape, dtype=torch.uint8),
  "reward": tensor(T, B),
  "done": tensor(T, B, dtype=torch.uint8),
  "policy_logits": tensor(T, B, num_actions),
  "baseline": tensor(T, B),
  "action": tensor(T, B, dtype=torch.int64),
}
\end{minted}
Here, \texttt{tensor(T, B)} is a tensor of shape \texttt{(T, B)},
where \texttt{T} is the unroll length, \texttt{B} is the batch size,
\texttt{obs\_shape} is a tuple of the observation shape (e.g.,
$[4, 84, 84]$ in the case of Atari with a frame stacking of the last
$4$ frames) and \texttt{num\_actions} is the number of discrete
actions of the environment.

In order to facilitate fast experiment turnaround times, the number of
actors should be large enough as to saturate the learner infeed, i.e.,
batches should be generated fast enough for the learner GPU to be fully
utilized. Speed is an important design goal of TorchBeast, but not its only
one. We will give an overview on our design principles in the next section.

\section{\TorchBeast{} Design Principles}
\label{sec:philosophy}

Ideally, researchers should be able to prototype their ideas quickly
without the mental overhead of low-level languages, but also without
the computational overhead of Python in places where it would
drastically impact performance.
There is a tension between those two goals.
Building frameworks with performance in mind can result in rigid constrains that
reduce how fast researchers can implement their ideas or even impair
their research directions.
%In our experience, the core assumptions of today are often
%the dogmas challenged tomorrow, in some cases precisely because they
%were hard to get around.
While TorchBeast necessarily relies on
engineering assumptions as well, we employ a few design principles
meant to offer researchers maximal leverage when implementing new ideas:

\paragraph{TorchBeast is not a framework.}  The TorchBeast repository
implements a certain type of agent and environment using the
IMPALA architecture. It is not meant to be imported as a
dependency but to be forked and modified in whatever way
necessary for a specific research goal. Compared to traditional
software engineering, the short half-life of research code makes this
approach more natural in the domain of deep reinforcement learning.

\paragraph{All machine learning code is in Python.}  Although the
PolyBeast variant of TorchBeast uses \Cpp\ components for its queuing
and batching logic, researchers should generally have no need to touch
those parts. In the special cases where they do, the changes necessary
should not involve digging through many layers of abstractions in the codebase.

\paragraph{One file to rule them all.}  While not strictly packaged as
a single file, TorchBeast tries to stay close to a ``one file only''
ideal. To take PolyBeast as an example, all agent code lives in
\texttt{polybeast.py} while the environment code lives in
\texttt{polybeast\_env.py}. No other files need to be touched to swap
out either the agent neural network model or the specific environment used
for training.

\subsection*{Adapting TorchBeast to research needs}

As a simple example of how to adapt PolyBeast to specific research needs,
we show pseudocode for changing the environment from Atari to the
MinAtar task suite~\cite{young19minatar}, a set of $10\times 10$ grid-world
Atari-like environments.
A user would fork TorchBeast, and modify the
following two parts only: (1) In \texttt{polybeast\_env.py}, change the
\texttt{create\_env} function to return a MinAtar environment, see
Figure~\ref{fig:minatar_in_tb_env}. (2) In \texttt{polybeast.py},
change the neural network model to use a smaller ConvNet, see
Figure~\ref{fig:minatar_in_tb}.

\begin{figure}[t!]
  \centering
  \begin{subfigure}[b]{0.49\textwidth}
    \centering\footnotesize
    \begin{minted}{python}
def create_env(flags):
  return atari_wrappers.wrap_pytorch(
    atari_wrappers.wrap_deepmind(
    atari_wrappers.make_atari(flags.env),
    clip_rewards=False,
    frame_stack=True,
    scale=False))
    \end{minted}
    \caption{In default \texttt{polybeast\_env.py}}
  \end{subfigure}\hfill
  \begin{subfigure}[b]{0.46\textwidth}
    \centering\footnotesize
    \begin{minted}{python}
def create_env(flags):
  env = minatar.Environment(flags.env)
  env = MinAtarEnv(env)  # Gym wrapper.
  return env



    \end{minted}
    \caption{In MinAtar fork of \texttt{polybeast\_env.py}}
    \end{subfigure}
  \caption{Changing TorchBeast to use MinAtar: Changes in
    \texttt{polybeast\_env.py}}
  \label{fig:minatar_in_tb_env}
\end{figure}

\begin{figure}
  \centering\footnotesize
  \begin{minted}{python}
class MinAtarNet(nn.Module):
  def __init__(self, num_actions, use_rnn=False):
    # ...
    self.conv = nn.Conv2d(4, 16, kernel_size=3, stride=1)
    self.core = nn.Linear(num_linear_units, 128)
    self.policy = nn.Linear(128, num_actions)
    self.baseline = nn.Linear(128, 1)

  def forward(self, inputs, core_state=()):
    # ...
    return (action, policy_logits, baseline), core_state
  \end{minted}
  \caption{Changing TorchBeast to use MinAtar: New model in
    \texttt{polybeast.py}}
  \label{fig:minatar_in_tb}
\end{figure}

\paragraph{More complex changes} Some research directions have more
specific needs.
For example, when using TorchBeast to train RL agents for
network congestion control~\cite{mvfst-rl}, the roles of clients and
servers in TorchBeast
needed to be reversed due to technical limitations of the simulator
used as an environment. This was easily achieved by forking the
TorchBeast repository and modifying the ``actor pool'' logic. Another
example of a larger extension involving changing logic in \Cpp\ would be
moving the rollout logic from doing cross-episode batches to doing
padding (by guaranteeing that each batch contains data from at most
one episode, this makes using certain models such as attention
easier). However, while such changes are straight-forward to do, we
believe that most research needs do not fall into this category and
can readily use TorchBeast by changing the agent parameterization and
environment.

\section{Experiments}

We test TorchBeast on the classic Atari
suite~\cite{bellemare13arcade}. The
hyperparameters, network architecture, and optimization procedure are
taken from the IMPALA
paper~\cite[Table~G.1]{DBLP:journals/corr/abs-1802-01561}. We run
PolyBeast on a single Nvidia Quadro GP100 GPU using 25~CPU cores
for the~48 environments.
For the agent model, we use a version of the ``deep network''
without an LSTM from the IMPALA paper and train for 200~million frames
per environment and run
(corresponding to 50~million ``agent steps'' due to action
repetitions).

\begin{figure}
  \makebox[\textwidth][c]{\includegraphics[height=1.4\textwidth, keepaspectratio]{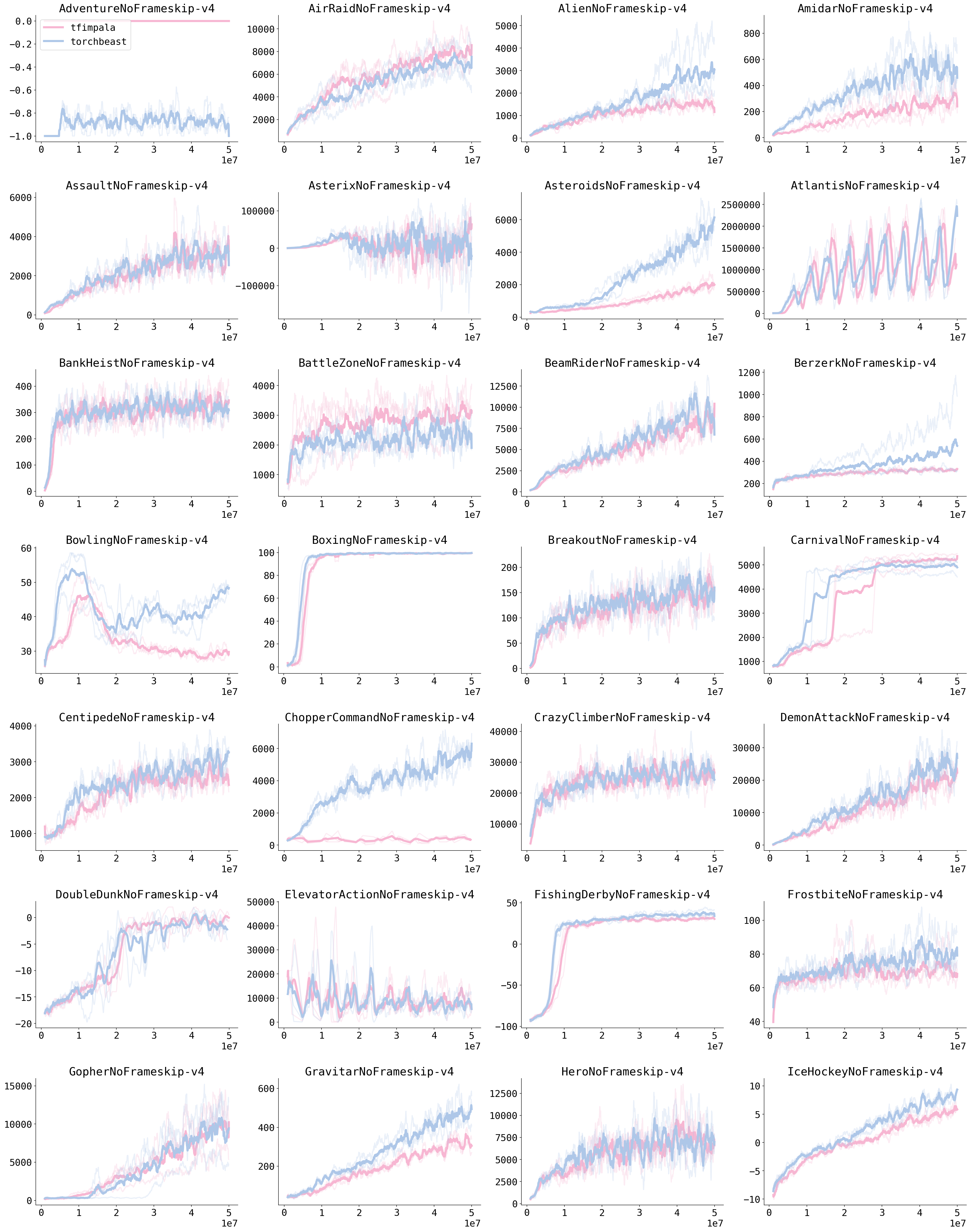}}
  \caption{TorchBeast (\textcolor{blue}{blue}) and TensorFlow IMPALA
    (\textcolor{red}{red}) runs on several Atari levels (first set).}
  \label{fig:atari_results_0}
\end{figure}

\begin{figure}
  \makebox[\textwidth][c]{\includegraphics[height=1.4\textwidth, keepaspectratio]{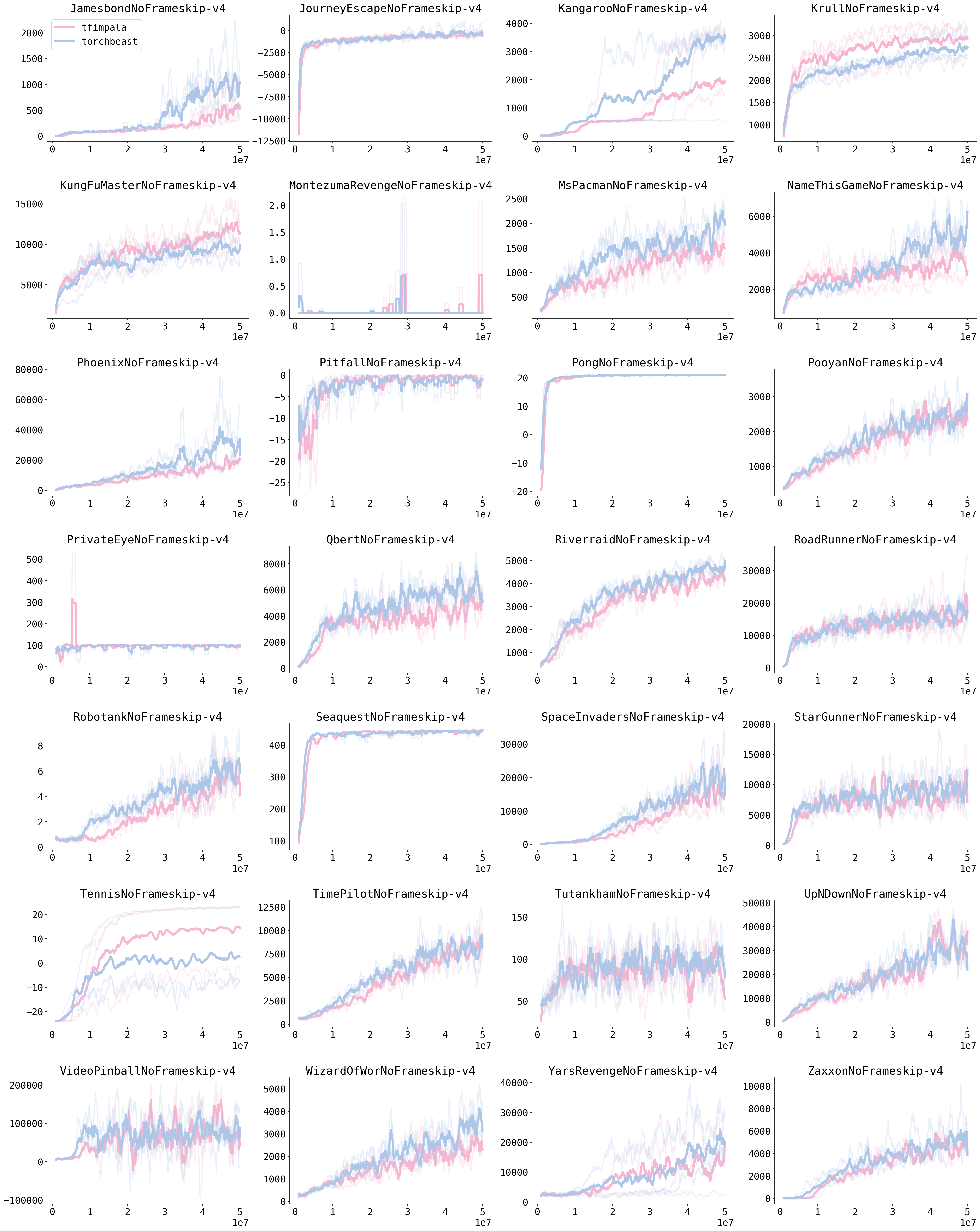}}
  \caption{TorchBeast (\textcolor{blue}{blue}) and TensorFlow IMPALA
    (\textcolor{red}{red}) runs on several Atari levels (second set).}
  \label{fig:atari_results_1}
\end{figure}

For the environments, we use the package and API made available in
OpenAI Gym~\cite{1606.01540}\footnote{We exclude the Atari game
  \emph{Defender} due to bugs in
  the Python~3.6 version of its OpenAI Gym environment; Python~3.6 is
  the latest Python version for which TensorFlow version 1.9.0, required
  by the open source version of TensorFlow IMPALA, is available.}
with a set of default ``environment wrappers'' provided by
OpenAI~\cite[see][file
  \nolinkurl{baselines/common/atari\_wrappers.py}]{openai/baselines}. These
wrappers provide preprocessing functionality common for doing RL on
Atari. They include action repetitions, frame stacking, frame warping
(resizing), frame ``max-pool-and-skip'', random no-ops at the
beginning of the game as well as functionality to end the RL episode on
life loss. We note that researchers often report total episodic
returns according to the original definition of episodes in Atari but
train with the end-of-life definition of episodes. This makes sense
when viewing the end-of-life episodes as a training detail but
also increases the complexity of the training and logging setup.
For simplicity, we train and report returns according to the same
end-of-life episode definition in our experiments. This reduces the
total episode returns reported by an average factor of the number of
lives in each Atari game.

We also took DeepMind's open source TensorFlow IMPALA implementation
from GitHub and modified it slightly to be compatible with Python~3 and
operate with the Gym API.\footnote{The slightly modified code can be
  found at \url{https://github.com/heiner/scalable_agent/releases/tag/gym}.}
The results of TorchBeast and TensorFlow IMPALA are reported in
Figures~\ref{fig:atari_results_0} and~\ref{fig:atari_results_1} and
demonstrate that both
implementations are on par for these tasks. We note that
the scores obtained by either implementation for many environments
fall short of the numbers reported
in~\cite[Table~C.1]{DBLP:journals/corr/abs-1802-01561}.
We believe
this is due to the question of episode definitions mentioned above and
possibly due to different environment preprocessing more
generally.\footnote{Anecdotally, even the resizing/downscaling method
  can impact the performance of RL agents.} Our results are equivalent
to DeepMind's IMPALA implementation when using the same preprocessing
and episode definitions.

Besides training performance, PolyBeast is also on par with
TensorFlow IMPALA when it comes to throughput (measured in
consumed frames per second) and thus experiment duration in wallclock
time.

\section{Implementation}
\label{sec:headings}

TorchBeast comes in two variants, dubbed ``MonoBeast'' and
``PolyBeast''. The main purpose of the MonoBeast variant is to be easy
to setup and get started with (no major dependencies besides Python
and PyTorch are required).
PolyBeast, on the other hand, utilizes Google's gRPC
library~\cite{grpc} for inter-process and transparent cross-machine
communication. It also implements the heavy-duty operations such as
batching as a Python extension module written in \Cpp. This allows us
to implement advanced features such as dynamic batching at the cost of
a more complex installation procedure. Either version uses multiple
processes to work around technical limitations of multithreaded Python
programs, see Section~\ref{subsec:gil} below for details.

\subsection{MonoBeast}

Since its earliest versions, PyTorch has support for
moving tensors to shared memory. In MonoBeast, we utilize this
feature in an algorithm that is roughly described as:

\begin{itemize}
  \item Create \texttt{num\_buffers} sets of \emph{rollout buffers},
    each of them containing shared-memory tensors without a batch
    dimension, e.g.,
\begin{minted}{python}
  buffers[0]['frame'] = torch.empty(T, *obs_shape, torch.uint8)
\end{minted}
  \item Create two shared queues, \texttt{free\_queue} and
    \texttt{full\_queue}. These queues will communicate integers using
    \UNIX\ pipes.
  \item Start \texttt{num\_actors} many \emph{actor processes}, each
    with a copy of the environment. Each actor dequeues an \texttt{index}
    from \texttt{free\_queue} and writes a batch-slice with rollout data into
    \texttt{buffers[index]}, then enqueues \texttt{index} to
    \texttt{full\_queue} and dequeues the next \texttt{index}.
  \item The main process has several \emph{learner threads}, each of
    which
    \begin{enumerate}
      \item  Dequeues \texttt{batch\_size}-many indices from
        \texttt{full\_queue}, stacks them together into a batch and
        moves them to the GPU, puts the indices back into
        \texttt{free\_queue}; then
      \item sends that batch through the model, computes losses, does
        a backward pass, and hogwild-updates the weights.
    \end{enumerate}
\end{itemize}

While simple, MonoBeast requires a relatively large amount of
constantly allocated shared memory, does model evaluations on the
actors on CPU instead of GPU, and involves a number of tensor copies not strictly
necessary. It is also limited to a single machine. To overcome
these downsides, we developed PolyBeast.

\subsection{PolyBeast}

There are two kinds of PolyBeast processes: A number of
\emph{environment servers} and a single \emph{learner process}.

Environment servers, once running, wait for incoming
gRPC connections and when a client learner process connects, create a
new copy of the environment to serve to the client while the
bidirectional streaming connection lasts.
In this bidirectional
stream, an environment server sends out observations, rewards and some
book-keeping data like a tensor indicating whether the current episode
ended. The client in turn responds with actions.
In our
implementation, the environment
servers load environments via a Python function that returns
environment objects compatible with the OpenAI Gym interface. In order
to not suffer from GIL contention (see section~\ref{subsec:gil}),
users should therefore limit the number of parallel connections per
server.

The learner process starts a number of actor threads (in \Cpp) to
connect to the environment servers. To facilitate ``inference''
evaluations of the received observations on the GPU, these
observations should be \emph{dynamically batched}. TorchBeast
implements a version of the dynamic batching functionality present in
DeepMind's IMPALA implementation\footnote{See
  \url{https://github.com/deepmind/scalable_agent/blob/master/batcher.cc}}:
each actor thread appends the environment output data to a queue, the
\emph{inference queue}. Another part of the system is responsible for
reading from this queue, evaluating a model (or otherwise creating a
minibatch of actions for a given minibatch of observations) and
setting the result. After \texttt{unroll\_length} many interactions,
the actor thread will concatenate the data and enqueue it to another
queue, the \emph{learner queue}. Another part of the system is
responsible for dequeing from this queue and updating the model based
on this batched rollout.

By using gRPC, PolyBeast transparently runs using either a
single-machine or a distributed setup. Distributing the
environments over several machines is necessary for large-scale
experiments with computationally costly games like StarCraft~II. For
more economical environments, the bottleneck tends to be on the agent
side (e.g., neural network evaluations, backward passes, or memory
constrains).

The parts of PolyBeast responsible for reading from the inference and
learner queues are precisely those involving machine learning logic
and are written in Python for easy accessibility by researchers.
In Python-like pseudocode, the PolyBeast agent process looks like
this:
\begin{minted}{python}
def main():
  model = Model()
  optimizer = Optimizer()

  inference_queue = DynamicBatcher(batch_dim=1)
  learner_queue = BatchingQueue(FLAGS.batch_size, batch_dim=1)
  actors = ActorPool(learner_queue, inference_queue,
                     FLAGS.unroll_length, FLAGS.server_addresses)

  inference_thread = threading.Thread(target=infer,
                                      args=(model, inference_queue))
  inference_thread.start()
  actors.run()  # Starts threads connecting to environments, fills queues.

  for env_outputs, actor_outputs in learner_queue:
    learner_outputs = model(env_outputs)
    loss = compute_loss(learner_outputs, actor_outputs, env_outputs)
    loss.backward()
    optimizer.step()
    print("One gradient descent step, loss was", loss)

    if learning_done():
      break

  actors.stop()
  inference_queue.close()
  learner_queue.close()
  inference_thread.join()

def infer(model, inference_queue):
  for batch in inference_queue:
    env_outputs = batch.get_inputs()
    actor_outputs = model(env_outputs)
    batch.set_outputs(actor_outputs)

def compute_loss(learner_outputs, actor_outputs, env_outputs):
  ...  # Compute vtrace (or some other loss)

main()  # Start program.
\end{minted}

\subsection{A Note on Python's Global Interpreter Lock}
\label{subsec:gil}

The engineering decisions around TorchBeast and many other parallel RL
architecture are heavily influenced by an implementation detail of
CPython, the Python programming language's widely used reference
implementation. In CPython, the global interpreter lock (``GIL'') is a
mutex protecting access to Python objects. Any thread executing Python
bytecode needs to hold the GIL, hence only a single thread will do so
at any point in time. This has been a core part of CPython's design
ever since its first multithreading support in 1997.  While
suggestions have been made to change CPython's design in this regard,
none have proved successful or seem likely to be adopted in the
future.

Since the development of asynchronous actor-critic
methods~\cite{DBLP:journals/corr/MnihBMGLHSK16}, running multiple
environments in parallel has been the norm in scalable RL
architectures and Python implementations of any such algorithm had to
contend with the GIL. Any naive multithreaded implementation of
asynchronous RL methods will fail to scale beyond more than a few
parallel environments. Since Python is the most popular language for
machine learning by a wide margin, several workarounds have been
proposed and used by the community. An obvious candidate is to use
separate processes and inter-process communication based on
\UNIX\ sockets, network communication, shared file handles or
\texttt{/dev/shm}-like shared-memory buffers. Another possibility is
to use multiple Python sub-interpreters in a single process.
Another approach is to in essence leave Python and move to
``in-graph'' environments in a system like TensorFlow either via
explicitly defined graphs or jit-compiled operations.
From an engineering perspective, all of these workarounds constitute a
sizable increase in complexity as well as a likely drop in efficiency
due to an increased number of copy operations and other overhead.

In the case of TorchBeast, we opted for multi-processing using (1)
sockets and shared-memory for MonoBeast, and (2) RPC calls and
\Cpp\ threads for PolyBeast.

\section{Conclusion}

We open-source TorchBeast, a platform for reinforcement learning research
implementing the popular IMPALA agent. Our design aims are to be
simple, fast and amenable to new research needs. We provide  an
implementation in plain Python, MonoBeast, as well as a
high-performance version, PolyBeast, capable of large-scale experiments
across several machines. In either version, all machine learning parts
are implemented in Python using PyTorch.

We evaluated TorchBeast on the Atari task suite and compared its
performance to the TensorFlow IMPALA implementation published by
its authors. When using the same environment preprocessing, both
implementations achieve equivalent performance in terms
of throughput, data efficiency, stability, as well as final
performance.

We discussed some engineering aspects for reinforcement learning
agents as well as examples of particular research directions and the
modifications of TorchBeast necessary to pursue them. We believe
TorchBeast provides a promising basis for reinforcement learning
research without the rigidity of static frameworks or complex
libraries.

\subsubsection*{Acknowledgments}

We would like to thank Soumith Chintala, Joe Spisak and the whole PyTorch
team for their support, and Alexander Miller, Pierre-Emmanuel Mazaré,
Roberta Raileanu, Victor Yuan Zhong and especially Jeremy Reizenstein
for their comments and insightful discussions.

\bibliographystyle{unsrt}

\bibliography{references}

% \appendix
% \section{Appendix}

\end{document}